\documentclass[lettersize,journal]{IEEEtran}
\usepackage{amsmath,amsfonts}
\usepackage{algorithmic}
\usepackage{algorithm}
\usepackage{array}
\usepackage[caption=false,font=normalsize,labelfont=sf,textfont=sf]{subfig}
\usepackage{textcomp}
\usepackage{stfloats}
\usepackage{url}
\usepackage{verbatim}
\usepackage{graphicx}
\usepackage{cite}
\usepackage{multirow}
\usepackage{pifont} 

\begin{document}

\title{State-of-the-art optical-based physical adversarial attacks for deep learning computer vision systems}

\author{Junbin Fang, You Jiang, Canjian Jiang, Zoe L. Jiang, Siu-Ming Yiu, Chuanyi Liu
\thanks{Zoe L. Jiang is the corresponding author.

Junbin Fang is with the Department of Optoelectronic Engineering, Jinan University, Guangzhou, 510632, China, and also with the Guangdong  Provincial  Key  Laboratory  of  Optical  Fiber  Sensing  and  Communications, Guangzhou, 510632, China, and also with the Guangdong Provincial Engineering Technology Research Center on Visible Light Commu-nication, and Guangzhou Municipal Key Laboratory of Engineering Technology on VisibleLight Communication, Guangzhou, 510632, China (e-mail: tjunbinfang@jnu.edu.cn). 

You Jiang and Canjian Jiang are with the Department of Optoelectronic Engineering, Jinan University, Guangzhou, 510632, China (e-mail: youjiang@stu2022.jnu.edu.cn; canjianjiang@foxmail.com). 

Zoe L. Jiang is with the School of Computer Science and Technology, Harbin Institute of Technology, Shenzhen, Shenzhen, 518055, China, and also with the Guangdong Provincial Key Laboratory of Novel Security Intelligence Technologies, Guangdong, 510632, China, and also with the Peng Cheng Laboratory, Shenzhen, 518055, China (e-mail: zoeljiang@hit.edu.cn).

Siu-Ming Yiu is with the Department of Computer Science, The University of Hong Kong, Hong Kong,  999077, China (e-mail: smyiu@cs.hku.hk).

Chuanyi Liu is with the School of Computer Science and Technology, Harbin Institute of Technology (Shenzhen), Shenzhen, 518055, China, and also with the Guangdong Provincial Key Laboratory of Novel Security Intelligence Technologies, Harbin Institute of Technology (Shenzhen), Shenzhen, 518055, China, and also with the Shenzhen Key Laboratory of Data Security, Harbin Institute of Technology (Shenzhen), Shenzhen, 518055, China, and also with the Peng Cheng Laboratory, Shenzhen, 518000, China (e-mail: liuchuanyi@hit.edu.cn). 
}
}

\markboth{Journal of \LaTeX\ Class Files,~Vol.~14, No.~8, August~2021}%
{Shell \MakeLowercase{\textit{et al.}}: A Sample Article Using IEEEtran.cls for IEEE Journals}


\maketitle

\begin{abstract}
 Adversarial attacks can mislead deep learning models to make false predictions by implanting small perturbations to the original input that are imperceptible to the human eye, which poses a huge security threat to the computer vision systems based on deep learning. Physical adversarial attacks, which is more realistic, as the perturbation is introduced to the input before it is being captured and converted to a binary image inside the vision system, when compared to digital adversarial attacks. In this paper, we focus on physical adversarial attacks and further classify them into invasive and non-invasive. Optical-based physical adversarial attack techniques (e.g. using light irradiation) belong to the non-invasive category. As  the perturbations can be easily ignored by humans as the perturbations are very similar to the effects generated by a natural environment in the real world. They are highly invisibility and executable and can pose a significant or even lethal threats to real systems. This paper focuses on optical-based physical adversarial attack techniques for computer vision systems, with emphasis on the introduction and discussion of optical-based physical adversarial attack techniques. 
\end{abstract}


\section{Introduction}
\IEEEPARstart{C}omputer vision uses computers and related equipment to simulate biological vision, so that computers have the ability to recognize three-dimensional environmental information through two-dimensional images. As an important part of the field of artificial intelligence (AI), in recent years, with the rapid development of deep learning technology, computer vision technology based on the deep neural networks (DNNs) has also made rapid progress and has a wide range of applications in face recognition \cite{bib1}, automatic driving \cite{bib2}, intelligent manufacturing \cite{bib3}, biomedicine \cite{bib4}, etc.

However, DNNs have some inherent problems, in particular, the unexplainability due to the black box nature of the algorithms, which make it difficult to obtain the causal factors associated with the essential characteristics of examples and the predicted results. These problems lead to some security loopholes in computer vision technology based on DNNs, which can be used by malicious attackers to launch adversarial attacks. Adversarial attack misleads the classifier to generate false predictions by making small perturbations to the original input. These small perturbations are imperceptible to humans, but can make the classifier generate false predictions with higher confidence. As shown in the literature \cite{bib5}, an adversarial attack against a traffic sign recognition system can cause the self-driving car to misclassify the traffic signs and thus cause the self-driving system to issue false control commands, which can lead to traffic accidents and even endanger the personal safety of the driver and passengers.

Adversarial attacks pose a huge challenge to the security of computer vision technology based on DNNs and even AI technology. At present, adversarial attacks are mainly divided into two categories: digital and physical adversarial attacks \cite{bib6} which assume that attackers can directly operate the input binary images at the pixel level. Digital adversarial attacks require direct access/modification of digital image data, which has a much lower feasibility in the real world and is difficult to generate effective real threats to computer vision systems. Unlike digital adversarial attacks, physical adversarial attacks physically implant perturbations to modify the input before it is being captured by the computer vision system and converted to a binary image (e.g. drawing markers on a real road sign). Physical adversarial attacks may not be  as effective as digital adversarial attacks as the perceptions of the perturbation cannot be controlled directly as for digital adversarial attacks. On the other hand, they can be implemented in the real world and have a significant or even fatal impact on real systems, so they are receiving more and more attention and research in recent years \cite{bib7,bib8,bib9}.

Since the receiving device (e.g. a camera) of a computer vision system is essentially a photoelectric sensor, the interference from optics can often have critical effects on the computer vision system; for example, the first death in an autonomous car occurred when a Tesla Model S collided with a white tractor trailer. Tesla stated the incident occurred because “neither Autopilot nor the driver noticed the white side of the tractor trailer against a brightly lit sky.” \cite{bib10}. Physical adversarial attack methods based on optical means have become a popular research direction in recent years, and some research results have been published so far \cite{bib11,bib12}. Based on a brief introduction of some basic concepts of adversarial attacks, this paper classifies physical adversarial attack techniques in terms of both invasive attacks and non-invasive attacks, and focus on optical-based physical adversarial attack techniques for computer vision systems, discuss and analyze in depth their features, mechanisms, and superiority (e.g., high invisibility, simple deployment).

In this paper, the analysis and research of physical adversarial attacks are helpful to the discovery of endogenous security loopholes of DNNs in advance and inspire the design of targeted defense solutions to reduce the security risks brought by the immaturity of AI and malicious applications, enhance AI security, and promote the deep application of AI.

\section{Some basic concepts of adversarial attacks}

In this section, we mainly introduce some basic concepts of adversarial attacks, first introduce the decision conditions and general classification of adversarial attacks, then analyze the causes of adversarial perturbations, and finally introduce the two main categories of adversarial attacks: digital adversarial attacks and physical adversarial attacks.
\subsection{Judgment conditions and classification of adversarial attacks}
\begin{figure}[htpb]
	\centering
	\includegraphics[width=3.3in]{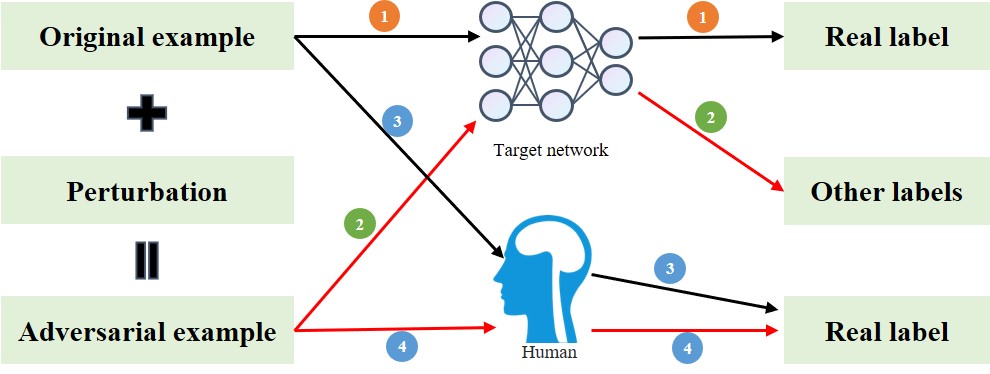}
	\caption{Conditions satisfied by the adversarial example.}
	\label{Fig1}
\end{figure}

Adversarial attacks were first proposed by Szegedy et al. \cite{bib23}, where a small perturbation of the original image can trick the target model making false predictions, where the image to which the perturbation has been added is called an adversarial example. However, with the development of adversarial attack techniques, adversarial examples can be either generated directly on a binary image (in the digital setting: by adding noise) or generated by capturing an attacked physical scene via image sensors such as cameras (in the physical setting: by adding fog, and sunlight). In general, an adversarial attack satisfies the following three conditions:
\begin{enumerate}
	\item{The target model can make a correct prediction of the original image before a perturbation is added to it, which is shown by ordinal number 1 in Fig.\ref{Fig1}.}
	\item{The target model must make a false prediction of the original image after a perturbation has been added to it, which is shown by ordinal number 2 in Fig.\ref{Fig1}.}
	\item{Before and after adding perturbations to the original image, humans can make correct predictions based on their perceptions, which is shown in Fig.\ref{Fig1} for ordinal number 3 (before adding perturbations) and ordinal number 4 (after adding perturbations).}
\end{enumerate}

Adversarial attacks can be divided into two categories based on the predicted outcome: non-targeted attack and targeted attack:

\subsubsection*{\bf Non-targeted attack}
 The attacker simply misleads the target model to output arbitrarily false predictions through the adversarial examples. It includes confidence reduction: the target model predicts a "dog" adversarial example as a "dog" image with a low confidence; Another is misclassification: the target depth model takes a "dog" adversarial example and predicts it as a "non-dog" image.
\subsubsection*{\bf Targeted attack}
The attacker tries to mislead the target model to output a specific false prediction by the adversarial examples. It includes targeted misclassification: the target model predicts any class of adversarial examples (e.g., "horse", and "cat" images) to a specific class (e.g., "dog" images). "Another is source/target misclassification: The targeted deep model takes an adversarial example of a specific class (e.g., "cat" images) and predicts it to another specific class (e.g., "dog" images). " class).

According to the time of the attack, the adversarial attack can also be divided into two categories: attack in the training stage and attack in the testing stage.

\begin{figure}[htpb]
	\centering
	\includegraphics[width=3.3in]{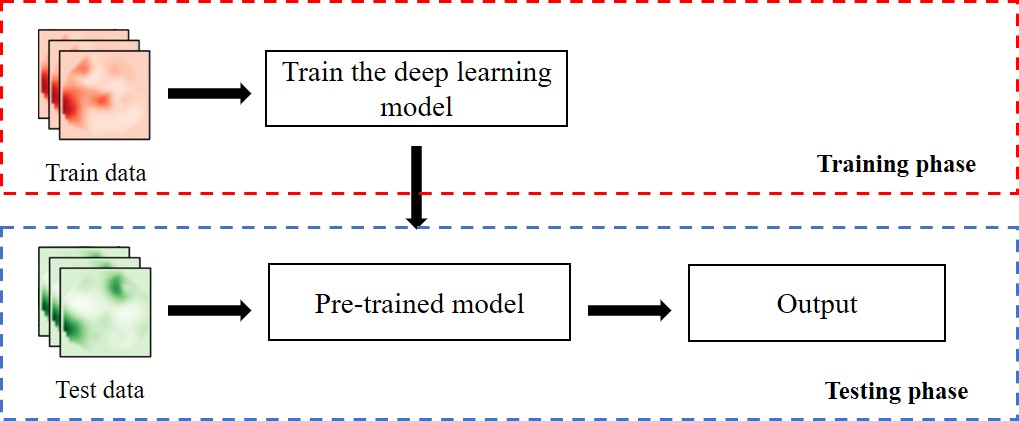}
	\caption{Deep learning process.}
	\label{Fig2}
\end{figure}

In general, the flow of a deep learning (DL) model includes two phases: the training phase (red dashed box) and the testing phase (blue dashed box), which is shown in Fig.\ref{Fig2}. Thus, the adversarial attack can also generate perturbations at different stages/times, including: training-phase attack (a.k.a. causative or poisoning attack) and testing-phase attack (a.k.a. exploratory or evasion attack) \cite{bib24}.

\subsubsection*{\bf Training phase attack}
In the training phase, the attacker can modify the training dataset by data injection, modifying data features and labels, etc. The aim is to make the target model learn the false target features, to interfere with and destroy the logic of the target model.

\subsubsection*{\bf Testing phase attack}
In the testing phase, the attacker does not and cannot interfere or destroy the logic of the target model, but the attacker can make adversarial examples based on the amount of knowledge of the target model obtained and use the adversarial examples to force the pre-trained target model to generate false predictions \cite{bib25}. According to the amount of knowledge of the target model obtained by the attacker, the test phase attacks can be further classified into: white-box attacks, black-box attacks, and gray-box attacks.

\begin{itemize}
	\item{White-box attack: The attacker assumes the complete knowledge of the targeted model, including its parameter values, architecture, training method, and in some cases its training data as well \cite{bib26}. Therefore, attackers can craft an adversarial example by obtaining the full amount of knowledge of the target's deep model.}
	\item{Black-box attack: The attacker cannot access the DNN model, and thus cannot obtain the model structure and parameters, and can only obtain the output result of the target model by inputting the original data to the target model \cite{bib27}.}
	\item{Gray-box attack: The attacker assumes partial knowledge of the target model \cite{bib28}. Therefore, attackers can use the known partial knowledge to obtain more unknown knowledge of the target model and reason out the vulnerability of the target model.}
\end{itemize}

\subsection{Reasons for adversarial perturbations}
Adversarial perturbations can misguide targeting objects by poisoning clean inputs to change their intrinsic structures learned by DNNs for recognition or classification \cite{bib29}. In recent years, a large number of researchers have tried to analyze the causes of adversarial perturbations from different perspectives, and although there is no unified statement on the causes of adversarial perturbations, in general, they can be divided into three categories of possible causes:

{\bf (1) Local over linearization of DNN structure, minimal perturbations may be amplified in the process of transmission.}

\begin{figure*}[htpb]
	\centering
	\includegraphics[width=4.5in]{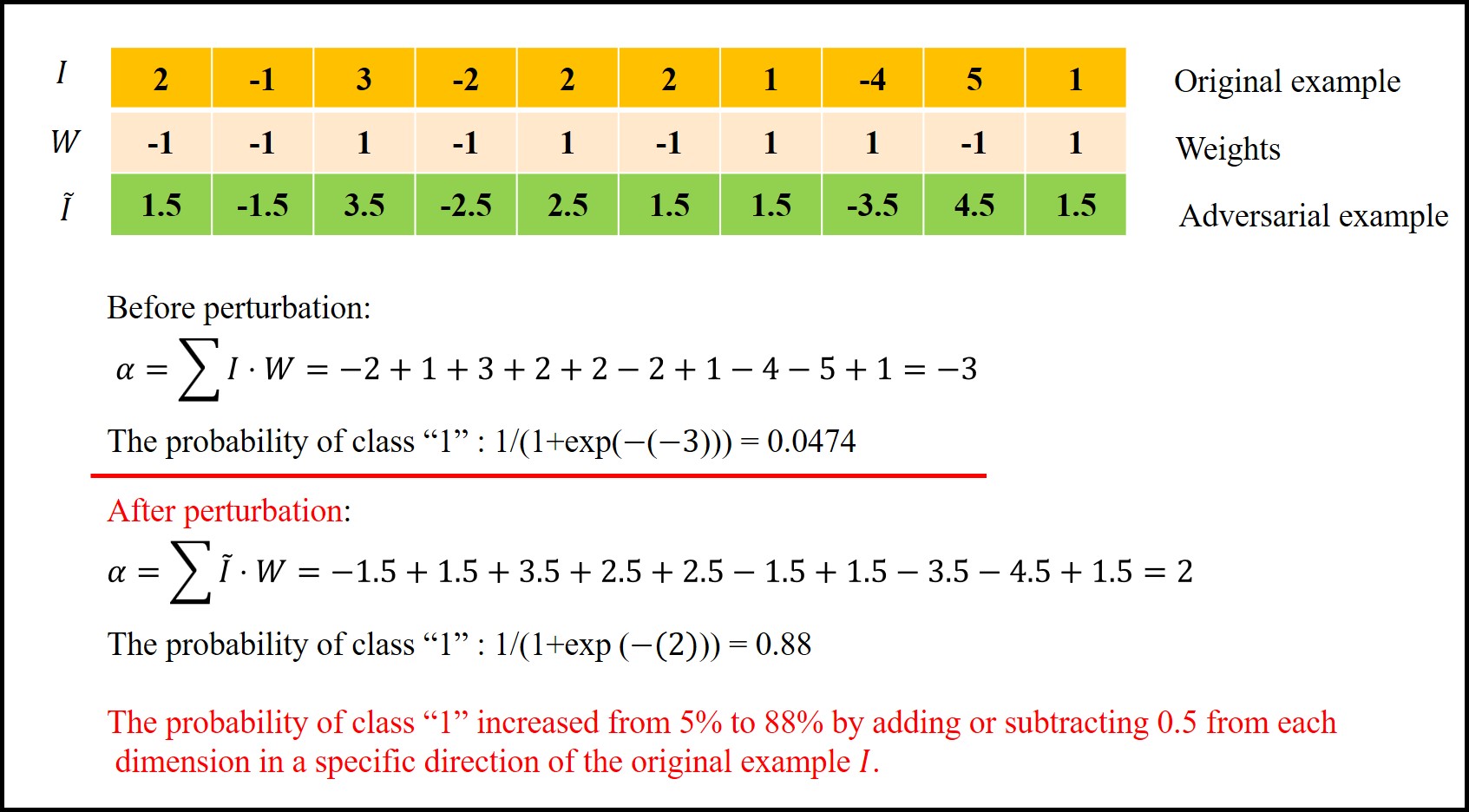}
	\caption{Minimal perturbations are scaled up in the high-dimensional linear classifier. Before the perturbation, the classifier classified the original image into class "1" with a probability of 5\%; by adding or subtracting 0.5 to each pixel of the original image to obtain the adversarial example , the classifier classified the adversarial example into class "1 "class with 88\% confidence (Image Credit: Zhang et al. \cite{bib31}).}
	\label{Fig3}
\end{figure*}

Goodfellow et al. \cite{bib30} believe that the cause of adversarial examples is the linear behavior in a high dimensional space. In the high-dimensional linear classifier, the minimal perturbations in each dimension are accumulated and amplified by dot product operations, and when the nonlinear activation function sigmoid is used to calculate the classification probability, the probability of the original image I being classified into "1" class before and after perturbation is increased from 5\% to 88\%, which is shown in Fig.\ref{Fig3}. If the DNN uses a linear activation function, the original image is more susceptible to a larger difference in classification probability before and after perturbation.

{\bf (2) The training data set contains insufficient target features, which causes the decision boundary of DNN to stop prematurely.}
	
Some researchers also argue that the vulnerability of DNN models is caused by incomplete training, which is attributed to insufficient data \cite{bib29}. The large dimensionality of the dataset due to the diversity of the target features will lead to premature stopping and weak generalization of the decision boundaries of the DNN model when the training data is under-labeled.
	
\begin{figure}[htpb]
	\centering
	\includegraphics[width=3.3in]{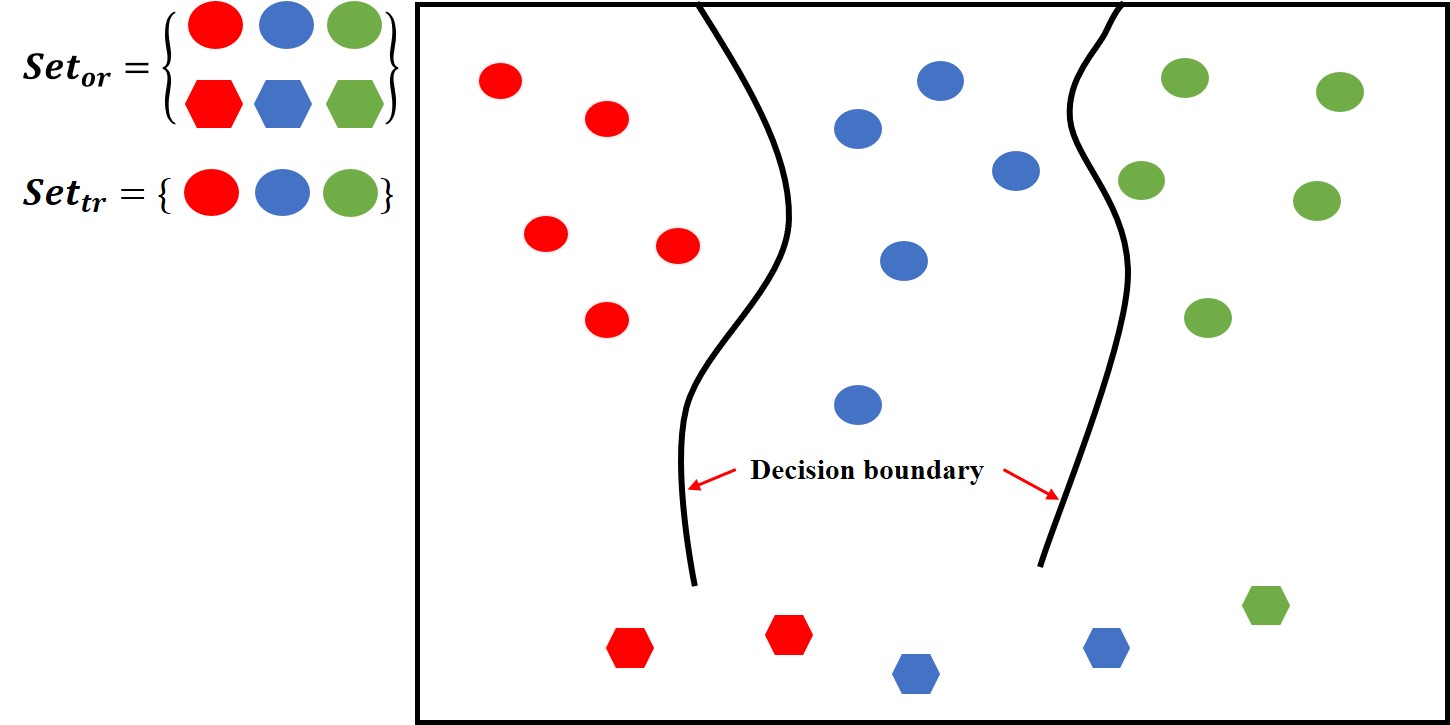}
	\caption{When the training dataset contains insufficient target features, it will cause the decision boundary of the target depth model to stop prematurely. Where ${Set}_{or}$ denotes the set of target features in the original dataset and ${Set}_{tr}$ denotes the set of target features in the training dataset.}
	\label{Fig4}
\end{figure}

Suppose that the original target dataset $Set_{or}$ has 6 features including red, blue, and green colors of circles and hexagons, but the randomly selected training dataset $Set_{tr}$ contains only red, blue, and green colors of circles features, which is shown in Fig.\ref{Fig4}. During the training process, when the target depth model can distinguish the features contained in $Set_{tr}$ well, the training may stop and the decision edge will also stop evolving. For some features not covered in the training dataset (e.g., hexagon), the target depth model will not be able to classify them correctly, so more data are needed for the target depth model to learn more target features to improve the robustness of the model.
	
{\bf (3) The presence of non-robust features in the classifier leads to inaccurate predictions.}
	
Cubuk et al. \cite{bib32} argue that the ‘‘origin of adversarial examples is primarily due to an inherent uncertainty that neural networks have about their predictions’’. They argue that the uncertainty in the output results of the target model is independent of the network architecture, training method, and data set. 

\begin{figure}[htpb]
	\centering
	\includegraphics[width=3.3in]{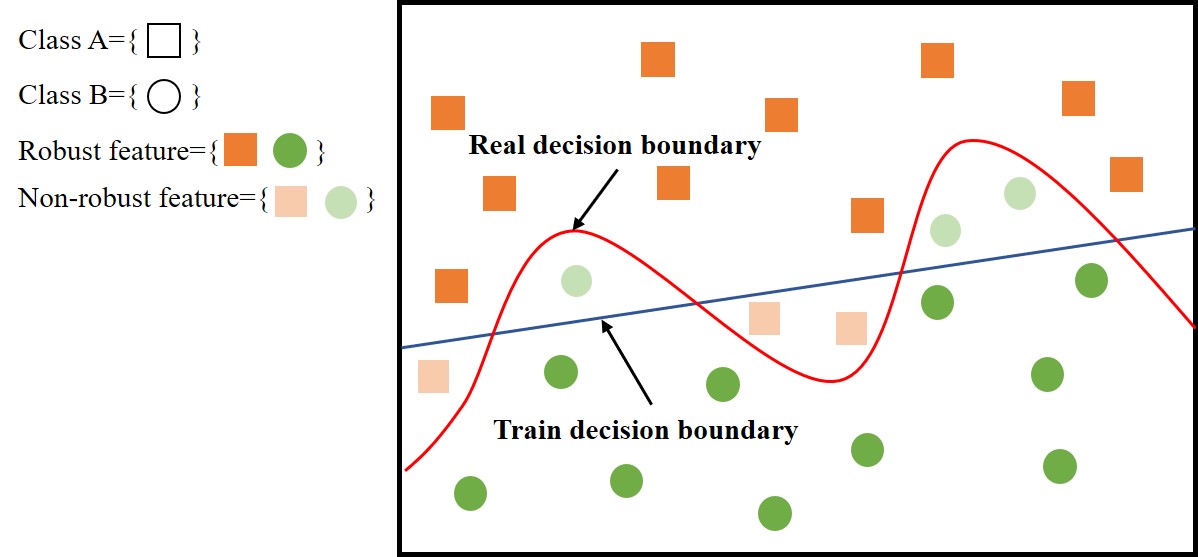}
	\caption{Non-robust features in the dataset are present in the classifier, resulting in inaccurate decision boundaries from the training.}
	\label{Fig5}
\end{figure}

Suppose that a classifier contains robust features (dark square and circle) and non-robust features (light square and circle), and the red curve is the real decision boundary of the dataset, which is shown in Fig.\ref{Fig5}. The presence of non-robust features in the classifier causes the decision boundary of the trained classifier (blue curve in Fig.\ref{Fig5}) to not recognize non-robust features in the target well, and therefore, if these non-robust features are added to the input image, it will likely result in the target depth model making false predictions.

\subsection{Digital adversarial attack VS physical adversarial attack}
\begin{figure*}[hbtp]
	\centering
	\includegraphics[width=7in]{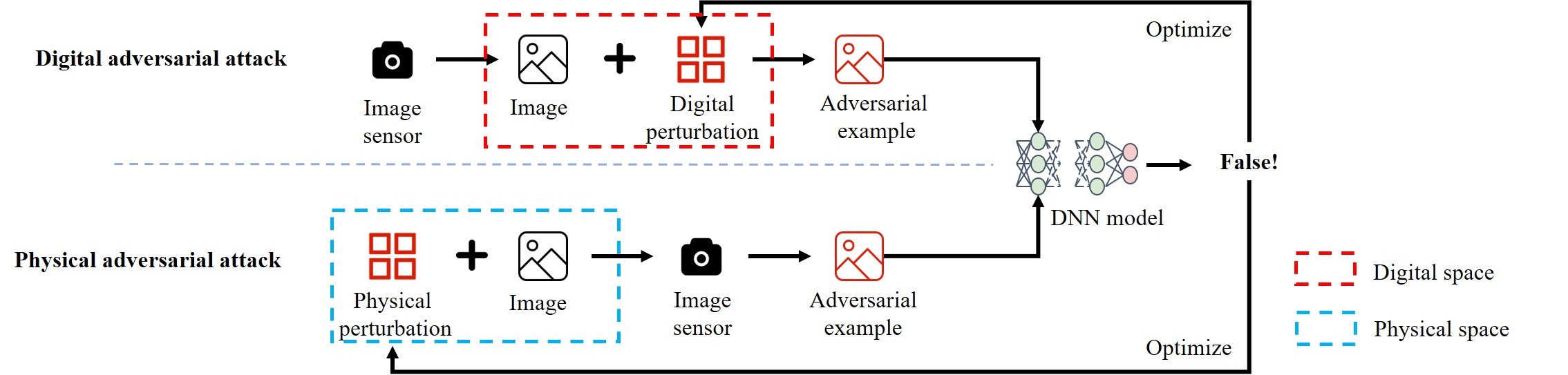}
	\caption{Comparison of digital adversarial attack and physical adversarial attack.}
	\label{Fig6}
\end{figure*}

In response to the large variety of approaches to adversarial attacks in recent years \cite{bib13,bib14,bib15,bib16,bib17,bib18,bib19,bib20}, Akhtar et al. \cite{bib33} try to use a simple/unified mathematical model to express the adversarial attacks as:
\begin{equation}
	M\left(\widetilde{I}\right)\rightarrow\widetilde{\ell}\ \ s.t.\ \ \widetilde{\ell}\neq\ell,\widetilde{I}\in S_I,\ M\left(I\sim\left\{S_I-\widetilde{I}\right\}\right)=\ell
\end{equation}

where $M\left(.\right)$ is the target model, i.e., $M\left(I\right):I\rightarrow\ell$, I is the original image and $\ell$ is the output of the target model. $\widetilde{I}=I+p$, for the adversarial example. As shown in Fig.\ref{Fig6}, if the attackers add perturbation to the image after the target is imaged, $\widetilde{I}$ is a digital  adversarial example; If the attacker attacks the image before the target is imaged, $\widetilde{I}$ is a physical adversarial example. Therefore, $S_I$ can be understood as the set of digital adversarial examples and physical adversarial examples, i.e., $\widetilde{I}\in S_I$.

Because the digital adversarial attack is to attack the image after the target is imaged, the attacker can use the program to produce pixel-level perturbation that is imperceptible to the human eye. Physical adversarial attacks require the capture of adversarial images by sensors (e.g., cameras), and since small perturbations are easily ignored when captured by cameras, the perturbations in most physical adversarial examples are easily detected by the human eye, but such perturbations do not appear to humans to affect the ensemble and thus do not raise alarms.

\subsubsection*{\bf Digital adversarial attack}
In a digital adversarial attack, the attacker first adds the perturbation $p$ directly to the original image $I$ through a program, and even that perturbation can be added precisely to a pixel point of $I$. Then, the size of $p$ is usually limited using the $l_n(n=0,1,\ 2,\infty)$ norm to ensure that the perturbation reaches imperceptibility to the human eye; finally, through the digital adversarial example $\widetilde{I}$ misled the target model to make false predictions. Along with these kinds of ideas, researchers have proposed many methods to generate optimal perturbations $p$. For example, Goodfellow et al. \cite{bib30} proposed fast gradient sign method $\left(FGSM\right)$ using $L_\infty$ norm constraint perturbation, Papernot et al. \cite{bib34} proposed jacobian-based saliency map attack $\left(JSMA\right)$ using $L_2$ norm constraint perturbation, Su et al. \cite{bib35} proposed One-Pixel attack $\left(OPA\right)$ using $L_0$ norm constraint perturbation, and Athalye A et al. \cite{bib36} proposed backward pass differentiable approximation attack $\left(BPDA\right)$ using $L_\infty$, $L_2$ norm constraints, and Carlini et al. \cite{bib37} proposed C\&W attack using $L_\infty$, $L_2$, $L_0$ norm constraints. Although digital adversarial attacks can exhibit high performance, these methods implant small perturbations by directly manipulating the input image at the pixel-level in digital world. Due to the influence of dynamic physical conditions (e.g., different shooting angles and distances, etc.) and optical imaging, it is difficult to perfectly transfer the perturbations in the digital adversarial examples to the real world, so the digital adversarial attack has very low executability in the real world.
\subsubsection*{\bf Physical adversarial attack}
In real applications, the camera is the entrance to the AI vision system, that is, the AI vision system needs to capture the original image through the camera as data input. Therefore, in the physical adversarial attacks, the attacker also needs to capture the adversarial example images through the system camera. However, since the imaging quality of the camera is limited by the physical performance of the camera's photoelectric sensor, small perturbations may not be captured by the camera, and in a way, it can also be argued that: small perturbations will be filtered out in the camera imaging stage. Therefore, in most existing physical adversarial attacks, the size of the implanted perturbation is usually within the perceptible range of the human eye, and the attack scheme often requires some additional means to hide the perturbation so that it is not noticed by the human eye.

\begin{figure}[hbtp]
	\centering
	\includegraphics[width=3.3in]{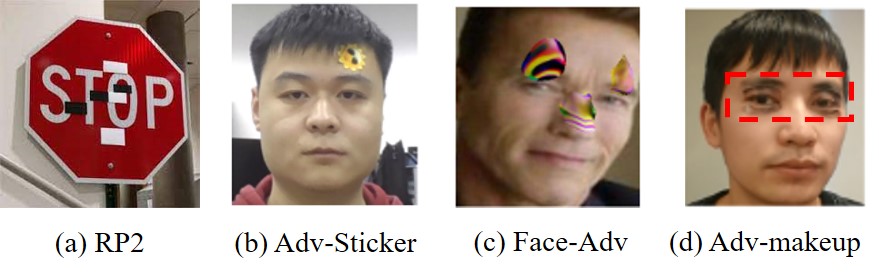}
	\caption{Some typical physical adversarial attacks. Among them (a), (b), (c), and (d) are the attack effect plots of RP2, Adv-Sticker, Face-Adv, and Adv-makeup, respectively. (Image(a) Credit: Eykholt K et al.\cite{bib7}, Image(b) Credit: Wei X et al.\cite{bib8}, Image(c) Credit: Shen M et al.\cite{bib9}, Image(d) Credit: Yin et al.\cite{bib39})}
	\label{Fig7}
\end{figure}

Kurakin et al. \cite{bib38} first found that the presence of physically adversarial examples, and proved that when the adversarial examples are printed out, they remain to be detrimental to the classifier even under different illumination and directions. Following this work, a large number of physical adversarial attack techniques have emerged. For example, Eykholt et al. \cite{bib7} proposed a generic attack algorithm—robust physical attack RP2 which finds the attack region of a traffic road sign using $L_1$ norm, then optimizes the attack region using $L_2$ norm, and finally manually attaches a black and white sticker to the optimal attack location. Although RP2 can capture robust perturbations in various environments, black-and-white stickers are not stealthy enough in the real world, which is shown in Fig.\ref{Fig7}(a), where black-and-white stickers are obvious on traffic road signs. To make the sticker attack more invisibility, Wei et al. \cite{bib8} proposed a real sticker attack $\left(Adv-sticker\right)$ technique, which is mainly used to find the optimal attack location by a region-based algorithm of differential evolution and 3D position transformation. Due to the use of sun stickers that are common in life, it will not arouse human suspicion on certain occasions, which is shown in Fig.\ref{Fig7}(b). To improve the diversity of stickers, Shen et al. \cite{bib9} proposed a FaceAdv using adversarial stickers, they designed an adversarial sticker production architecture consisting of a sticker generator and a converter, where the generator can generate multiple shapes of stickers, and the converter digitally pastes the stickers onto the human face and provides feedback to the generator for efficiency, the effect of the attack is shown in Fig.\ref{Fig7}(c) shows. To improve the invisibility of the perturbation, Yin et al. \cite{bib39} proposed a generic adversarial makeup attack $\left(Adv-makeup\right)$method to generate imperceptible eye shadows on the human eye region.

To improve the generalization capability of adversarial patches, Liu et al. \cite{bib40} exploited the perceptual and semantic biases of the model to generate category-independent universal adversarial patches. Huang et al. \cite{bib41} proposed a universal physical camouflage attack $\left(UPC\right)$ method that generates adversarial patches by jointly spoofing region proposal networks and misleading classifier and regression output errors, and can attack objects belonging to the same category of all objects. To solve the problem of performance degradation of the adversarial patch when the shooting viewpoint is shifted, Hu et al. \cite{bib42} proposed a toroidal-cropping-based expandable generative attack $\left(TC-EGA\right)$ to generate an adversarial texture with repetitive structure. The adversarial texture is then used to generate T-shirts, skirts, and dresses, etc., so that a person wearing such clothes can accomplish the " invisibility" effect on pedestrian detectors from different viewpoints.
\section{Optical adversarial attack techniques}

As shown in Fig.\ref{Fig7} (a), (b), and (c), the physical adversarial attacks all use non-transparent perturbations to overlay the target features, which are more prominent in the adversarial examples and easily attract human attention, and belong to the invasive attack. The main process of an invasive attack: a) the attacker first trains the target model to get the best attack position and optimal perturbation, and b) then sticks the non-transparent perturbation on the best attack position. However, in the real world, the non-transparent perturbations may be destroyed by the natural environment or human factors before the attack occurs.

\begin{table*}[hbtp]
	\begin{center}
		\caption{Optical-based Physical Adversarial Attack.}
		\label{tab1}
		\renewcommand{\arraystretch}{1.3}
		\begin{tabular}{| c | c | c | c | c |}
			\hline
			  & Method & Year & Physical tools for generating perturbation & imperceptible\\
			  \cline{2-5}	
			  \multirow{8}*{Light irradiation} & BulbAttack \cite{bib47} & 2021 & Bulb panel & \ding{56}  \\
			  \cline{2-5}			  
			  ~ & Invisible mask \cite{bib11} & 2018 & LED & \ding{56}  \\ 
			  \cline{2-5}
			  ~ & Pony et al. \cite{bib51} & 2021 & LED & \ding{56} \\ 
			  \cline{2-5}
			  ~ & Sun et al. \cite{bib49} & 2020 & Laser & \ding{56}  \\ 
			  \cline{2-5}
			  ~ & LightAttack \cite{bib52} & 2018 & Laser spot & \ding{56}  \\ 
			  \cline{2-5}
			  ~ & AdvLB \cite{bib12} & 2021 & Laser & \ding{56} \\ 
			  \cline{2-5}
			  ~ & AdvNB \cite{bib13} & 2022 & Neon beam & \ding{56}  \\
			  \cline{2-5}
			  ~ & ShadowAttack \cite{bib14} & 2022 & Shadow & \ding{56}  \\ 
			  \hline
			  \cline{2-5}			 	
				\multirow{14}{*}{Imaging device manipulation} & ISPAttack \cite{bib45} & 2021 & Camera & \ding{52}  \\ 
				\cline{2-5}
				~ & Adverial camera stickers \cite{bib15} & 2019 & Camera & \ding{56}  \\ 
				\cline{2-5}
				~ & TranslucentPatch \cite{bib46} & 2021 & Camera & \ding{56}  \\ 
				\cline{2-5}
				~ & LEDAttack \cite{bib43} & 2021 & Camera & \ding{52}  \\ 
				\cline{2-5}
				~ & AdvZL \cite{bib16} & 2022 & Camera & \ding{56}  \\
				\cline{2-5} 
				~ & ALPA \cite{bib17} & 2020 & Projector & \ding{56}  \\ 
				\cline{2-5}
				~ & SLAP \cite{bib44} & 2021 & Projector & \ding{56}  \\ 
				\cline{2-5}
				~ & ProjectorAttack \cite{bib50} & 2019 & Projector & \ding{56}  \\ 
				\cline{2-5}
				~ & SPAA \cite{bib48} & 2022 & Projector & \ding{52}  \\ 
				\cline{2-5}
				~ & OPAD \cite{bib18} & 2021 & Projector & \ding{52}  \\ 
				\cline{2-5}
				~ & Li et al. \cite{bib19} & 2022 & Structured light & \ding{56}  \\ 
				\cline{2-5}
				~ & She et al. \cite{bib20} & 2019 & Visible light & \ding{52}  \\ 
				\cline{2-5}
				~ & SLMAttack \cite{bib21} & 2022 & Phase modulation module  & \ding{52}  \\ 
				\cline{2-5}
				~ & Chen et al. \cite{bib22} & 2021 & LED & \ding{52}  \\
				\hline 
		\end{tabular}
	\end{center}
\end{table*}

Unlike invasive attacks, non-invasive attacks can hide the perturbation well into the target features without attracting human attention. Optical-based physical adversarial attacks are a good form of non-intrusive attacks, such as optical perturbations generated by using visible light or optical devices (e.g., cameras, and projectors). Since the perturbation caused by such perturbations are more common in life, humans rely on life experience to prompt the brain to selectively ignore them. According to the form of perturbation, we divide the existing optical-based adversarial attacks into based on light irradiation and imaging device manipulation, which is shown in Table.\ref{tab1}.

\begin{figure}[hbtp]
	\centering
	\includegraphics[width=3.3in]{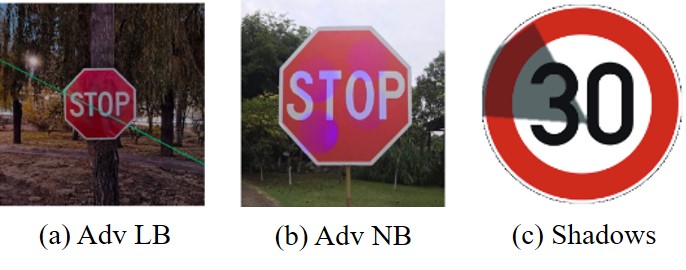}
	\caption{Optical physical adversarial attacks based on light irradiation. Where (a), (b) and (c) are the effect plots of Adv LB, Adv NB, and Shadows attacks, respectively. (Image(a) Credit: Duan et al. \cite{bib12}, Image(b) Credit: Hu et al. \cite{bib13}, Image(c) Credit: Zhong et al. \cite{bib14})}
	\label{Fig8}
\end{figure}

\subsection{Optical-based physical adversarial attacks based on light irradiation}
As mentioned earlier, non-transparent perturbations of invasive attacks are easily destroyed by the natural environment or human factors. To improve the robustness of perturbations against external factors, some methods for generating optical perturbations using light-emitting devices have been proposed \cite{bib1,bib12,bib13,bib14}. For example, Zhu et al. \cite{bib47} designed a multi-bulb panel to attack the infrared thermal imaging detector, Zhou et al. \cite{bib11} used infrared LEDs to deceive a face recognition system by projecting infrared illumination perturbations onto a human face. Although the infrared lighting perturbation is invisible to the human eye, the attack is difficult to deploy the infrared LEDs to the optimal position and the use of an infrared cutoff filter can effectively filter out these infrared perturbations. Pony et al. \cite{bib51} proposed a method to attack the action recognition system by controlling LED lights. Sun et al. \cite{bib49} proposed the first black-box spoofing attack against LPS. They manipulate the laser to simulate occlusion mode and sparse point clouds away from autonomous vehicle. Nichols et al. \cite{bib52} Project light points to specific areas of the target for attack. Duan et al. \cite{bib12} proposed an adversarial laser beam $\left(AdvLB\right)$, which mainly acts the laser beam as an optical perturbation to the original image, and since the perturbation is generated by the laser light source, it will alleviate the problem of difficult deployment of the irradiating device. However, a green laser beam runs through the whole image, and this optical perturbation is successfully added to the target background, which is shown in Fig.\ref{Fig8}(a). Hu et al. \cite{bib13} proposed adversarial neon beam $\left(AdvNB\right)$, which is shown in Fig.\ref{Fig8}(b). It can add only the neon beam to the target foreground, thus further improving the invisibility of optical perturbation. Although beam perturbation can alleviate the difficult of device deployment, the perturbation is generated by a human-made light source, and there are some differences with the optical perturbations generated by natural phenomena, so they still attract human attention in some ways. To address this problem, Zhong et al. \cite{bib14} proposed a physical adversarial attack generated by a natural phenomenon, the shadow attack, which is shown in Fig.\ref{Fig8}(c). The shadow attack makes full use of the adversary—shadow nature of illumination, and since shadows are more common in the real world, leading humans to selectively ignore shadow regions, this method not only hides shadow perturbations well in the real world but also reduces human alertness.

\begin{figure}[hbtp]
	\centering
	\includegraphics[width=3.3in]{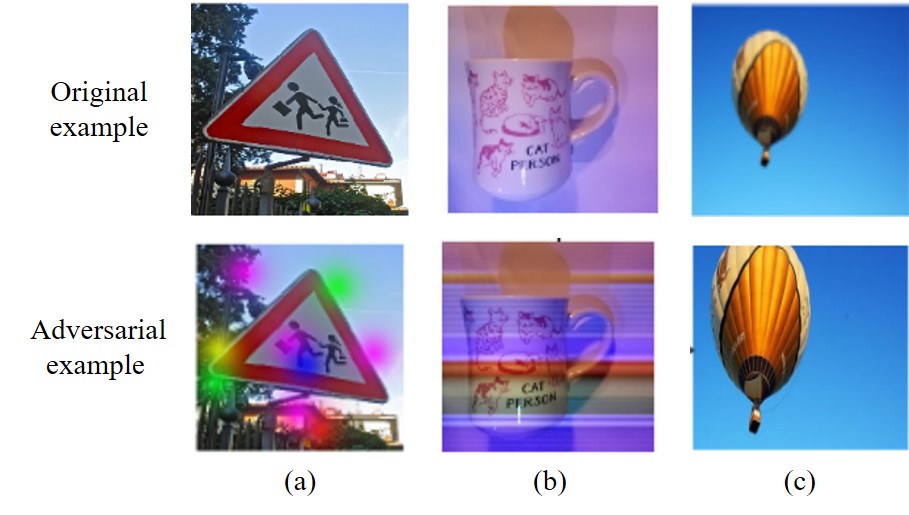}
	\caption{Physical adversarial attack against the camera, where (a) is an adversarial example generated by adding a translucent sticker to the lens, (b) is an adversarial example generated by the roll-up shutter effect, and (c) is an adversarial example generated by adjusting the camera's zoom lens. (Image(a) Credit: Li et al. \cite{bib15}, Image(b) Credit: Sayles A et al. \cite{bib43}, Image(c) Credit: Hu et al. \cite{bib16})}
	\label{Fig9}
\end{figure}

\begin{figure*}[hb]
	\centering
	\includegraphics[width=6in]{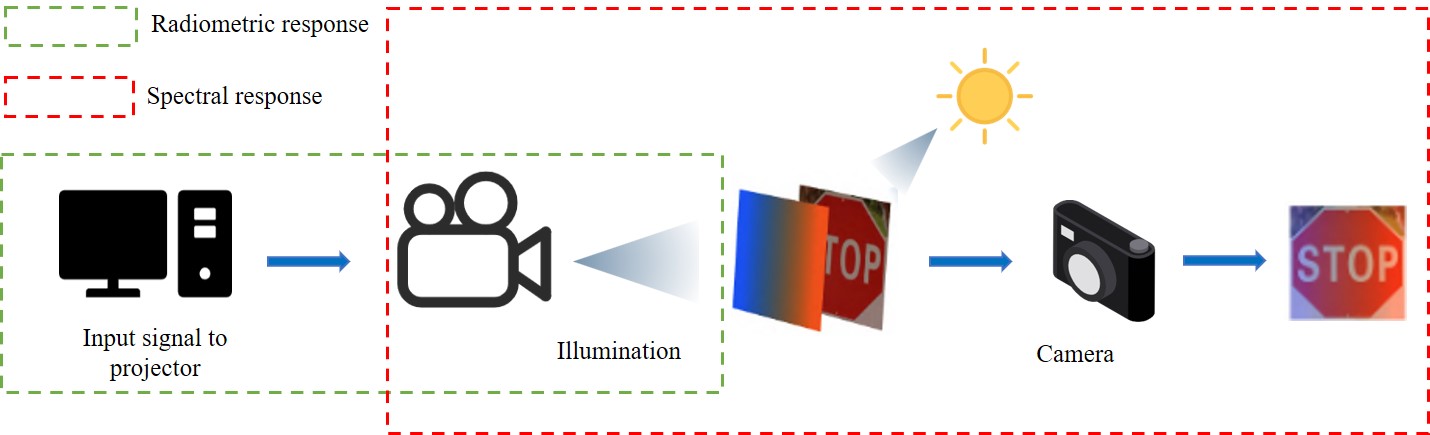}
	\caption{Optical adversarial attack $\left(OPAD\right)$.}
	\label{Fig10}
\end{figure*}

\subsection{Optical-based physical adversarial attacks based on imaging device manipulation}
Another novel non-intrusive optical-based physical adversarial attack is the physical manipulation of optical imaging devices to generate perturbations. For example, Phan et al. \cite{bib45} directly modify the RWA image captured by the sensor to deceive the image processing pipeline (ISP) hardware of the camera. Li et al. \cite{bib15} paste a carefully designed translucent sticker on the camera's lens, so at the moment of camera exposure, the perturbation on the sticker is projected onto the target and a physical adversarial example is generated, which is shown in Fig.\ref{Fig9}(a). Zolfi et al. \cite{bib46} successfully attacked the advanced driving assistance system (ADAS) of Tesla model X using the same attack method. Sayles A et al. \cite{bib43} used the rolling shutter effect to generate a perturbation that is imperceptible to the human eye but is captured by the camera, which is shown in Fig.\ref{Fig9}(b). Hu et al. \cite{bib16} proposed an adversarial zoom lens attack $\left(AdvZL\right)$ that scales the image by changing the zoom lens of the camera, achieving to successfully attack the target model without adding any perturbation to the original example for the first time, which is shown in Figure.\ref{Fig9}(c). These attacks against the camera, although simple to deploy, have a perceptible change in the texture and size of the original image, which would raise suspicion in the observer.

Another optical-based non-intrusive physical adversarial attack focuses on influencing AI vision systems by manipulating projection devices. For example, Nguyen et al. \cite{bib17} proposed a transformation-invariant perturbation generation method to perform real-time light projection attacks on face recognition systems. Lovisotto et al. \cite{bib44} use a projector to project the perturbation onto the target. Man et al. \cite{bib50} uses an electromagnetic signal generator, such as a light source, to transmit electromagnetic signals (light) to interfere with the signals (images) captured by the target sensor. Huang et al. \cite{bib48} projected the perturbation onto the target through a projector, and proposed a stealthy projectorbased adversarial attack (SPAA) method. Gnanasambandam et al. \cite{bib18} proposed a non-intrusive attack with structured illumination, which uses a low-cost projector-camera system to change the features of the target, and realizes an effective optical adversarial attack $\left(OPAD\right)$ against 3D objects, which is shown in Fig.\ref{Fig10}. OPAD consists of the Radiometric response of the projector (green dashed box) and the Spectral response of the camera projector (red dashed box). OPAD first uses the computer to generate perturbations; then uses the nonlinear mapping of the projector to convert the perturbation signal into an optical signal, finally, the optical perturbation is projected onto the image to change the texture feature of the target, and the camera is used to capture the modified image to generate adversarial examples. Li et al. \cite{bib19} used this idea by using a fringe projection imaging system $\left(FPP\right)$ and a 3D reconstruction algorithm to interfere with the face recognition system.

\begin{figure*}[htpb]
	\centering
	\includegraphics[width=6in]{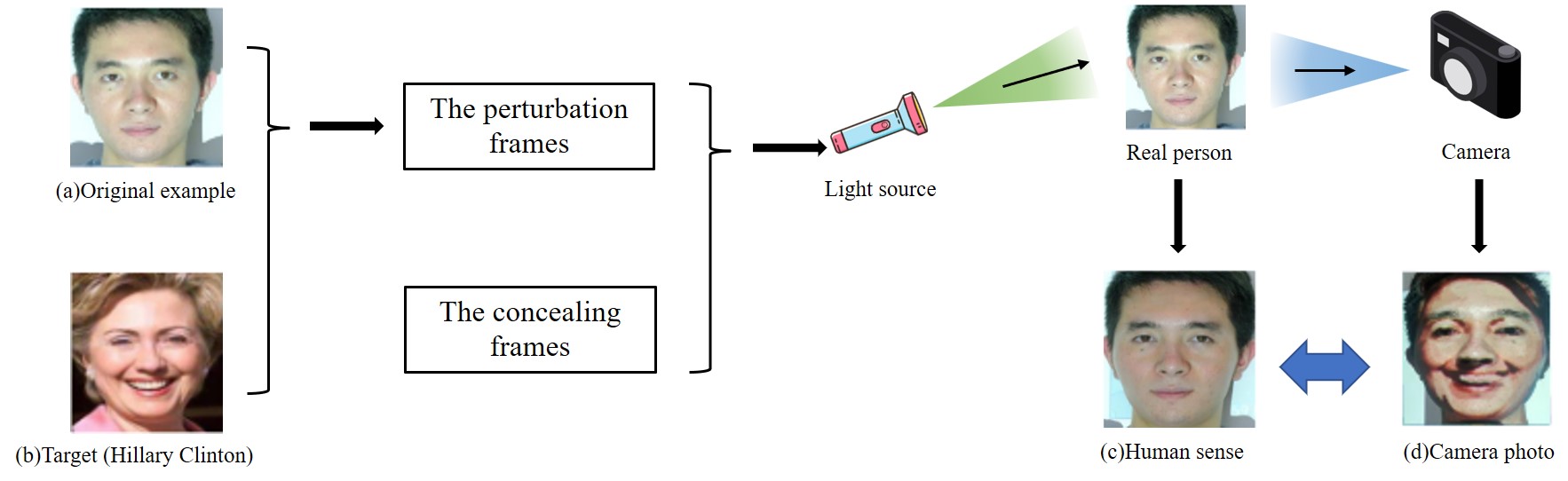}
	\caption{Visible light attack system. Where image (a) is the original image, image (b) is the target person, image (c) is an example of the confrontation observed by the human eye, and image (d) is an example of the confrontation captured by the camera.}
	\label{Fig11}
\end{figure*}

Although the above attack methods enhance the invisibility of perturbation in the real world, it has not yet reached a state that is completely imperceptibility to the human eye. In response to this problem, She et al. \cite{bib20} proposed a directed visible light attack using the Persistence of Vision $\left(POV\right)$ theory (i.e., when the conversion rate of light exceeds 25 Hz, the human brain will not be able to process these changes). It uses an alternate source consisting of a scrambled post and a hidden post to irradiate the human face, which is shown in Fig.\ref{Fig11}. The method achieves true imperceptibility to the human eye, which is shown in Fig.\ref{Fig11}(c). But the adversarial example obtained through the camera (Fig.\ref{Fig11}(d)) severely alters the texture features of the target.

\begin{figure*}[htpb]
	\centering
	\includegraphics[width=6in]{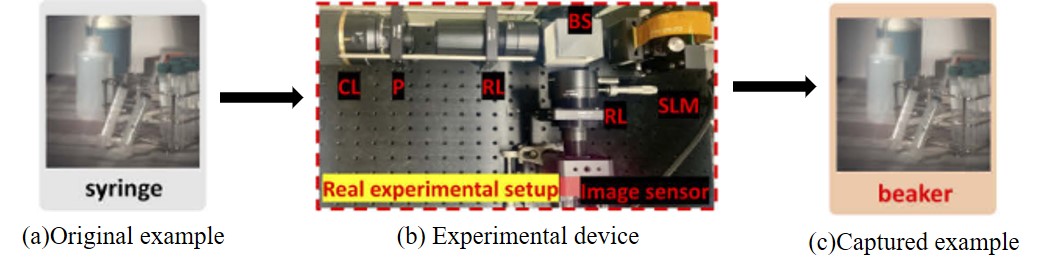}
	\caption{Engineering pupil function for optical adversarial attacks.}
	\label{Fig12}
\end{figure*}
While the visible light attack system generates imperceptible perturbation to the human eye, the adversarial examples obtained through the camera are severely distorted, which is shown in Fig.\ref{Fig11}(d). To be able to obtain fidelity of the adversarial examples using the camera, Kim et al. \cite{bib21} introduces a novel approach for non-targeted adversarial attack by modifying the light field information inside the optical imaging system. The idea is to modulate the phase information of the light in the pupil plane of an optical imaging system, which is shown in Fig.\ref{Fig12}. The authors designed a phase modulation module (Fig.\ref{Fig12}(b)) to generate the adversarial examples. A phase modulation module consisting of a polarizer (P), relay lens (RL), beam-splitter (BS), and spatial light modulator $\left(SLM\right)$ is implemented to the photography system. The method achieves perfect perturbation invisibility and is not distorted in the camera-captured adversarial examples, which is shown in Fig.\ref{Fig12}(c). However, it is difficult to simulate a qualified phase modulation module, so it is very difficult to deploy the experimental device.

\begin{figure*}[htpb]
	\centering
	\includegraphics[width=5in]{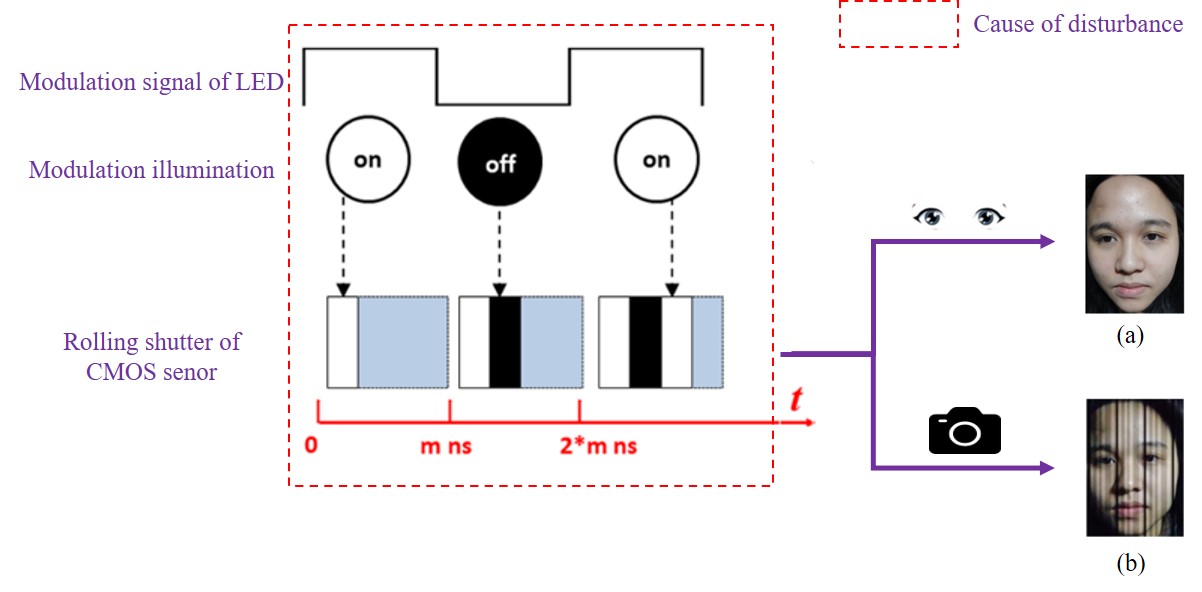}
	\caption{Novel attack scheme based on LED illumination modulation. Where, the red dashed box is the mechanism of imperceptible LED high-frequency flicker interacting with the CMOS roll-up shutter effect to generate bright and dark streaks for the human eye, (a) is the adversarial example observed by the human eye, and (b) is the adversarial example captured by the CMOS camera.}
	\label{Fig13}
\end{figure*}

Chen et al. \cite{bib22} propose a novel attack scheme based on LED illumination modulation, which not only generates perturbation imperceptible to the human eye in the real world but is also simple to deploy in the experimental setup. As can be seen in Fig.\ref{Fig13}(a), the perturbation generated based on the LED illumination modulation is imperceptible to the human eye, and as can be seen in Fig.\ref{Fig13}(b), the adversarial example captured by the camera contains a large number of black fringes, which are light perturbations generated by the interaction of the rolling shutter effect and modulated LED illumination in the CMOS camera imaging mechanism. When the LED illumination is on, bright pixels are stored in the active image column pixels; when the LED illumination is off, black edges are stored in the image column pixels due to underexposure of the camera, which is shown in the red dashed box in Fig.\ref{Fig13}. Because the "on/off" state of LED lighting changes at short intervals and the flicker frequency is beyond the limit of what the human eye can observe, the rapid flicker and bright /dark fringe perturbations caused by LED illumination modulation are not detectable to the human eye.

On this basis, the literature \cite{bib22} proposes two practical and novel attack methods for face recognition systems: denial-of-service $\left(DoS\right)$ attacks and escape attacks, the effects of which are shown in Fig.\ref{Fig14}.

\subsubsection*{\bf DoS Attacks}
\begin{figure}[hptb]
 	\centering
 	\includegraphics[width=2.8in]{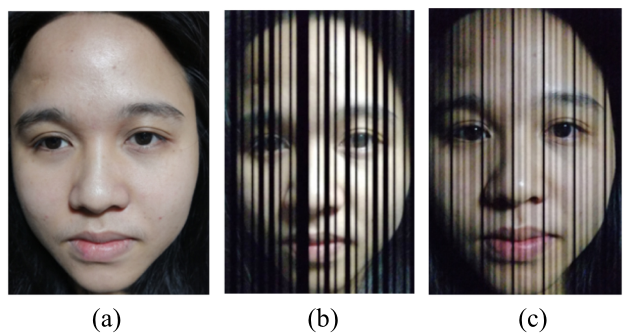}
 	\caption{Attack based on LED lighting modulation: (a) is the original image, (b) is an adversarial example supporting DoS attack, and (c) is an adversarial example supporting escape attack.}
 	\label{Fig14}
 \end{figure}

 Wide black fringes are generated by reducing the flicker frequency of the LED. Wide fringes may completely cover important facial features (such as eyes or nose), which is shown in Fig.\ref{Fig14}(b). Due to the lack of key facial features of the adversarial example, the face recognition system may be paralyzed in the target detection stage. Fig.\ref{Fig15}(a) verifies that as the distance of the DoS attack gets shorter, the higher the percentage of faces in the image, resulting in more wide black fringes on the face, and more of these black stripes covering key features of the face, ultimately resulting in a lower success rate of the face recognition system.
 
\subsubsection*{\bf Escape attacks}
\begin{figure}[hpbt]
	\centering
	\includegraphics[width=3.3in]{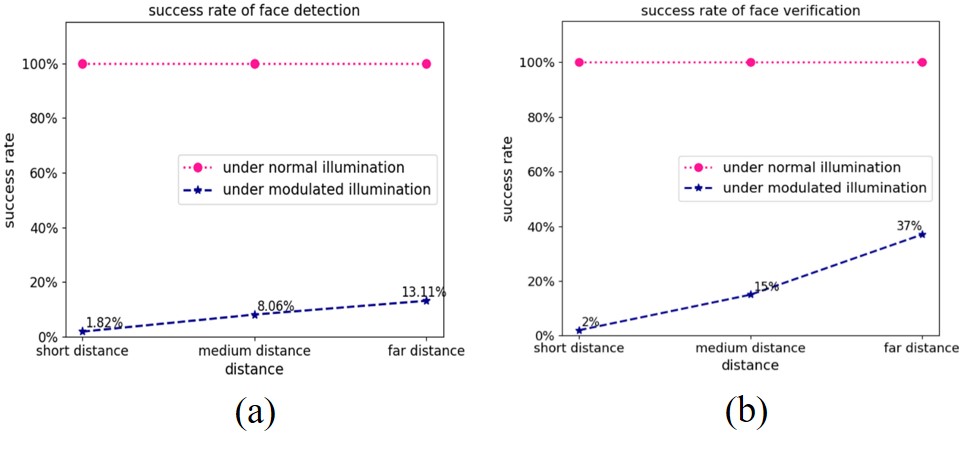}
	\caption{The success rate of face recognition system at different distances under normal illumination and illumination modulated by LED. Where (a) is the experimental result under DoS attack and (b) is the experimental result under escape attack.}
	\label{Fig15}
\end{figure}

Narrower black fringes are generated by accelerating the flicker frequency of the LED, which is shown in Fig.\ref{Fig14}(c). More narrow fringes not only add a large amount of repetitive and useless gradient information but also cover up many fine facial features, resulting in less difference between two different faces, which may eventually cause the face recognition system to judge two different faces as the same person. Fig.\ref{Fig15}(b) verifies that the shorter the distance of the escape attack, the narrower the black fringes cover the face features and also obscure a large number of fine facial features, ultimately leading to a lower success rate of the face recognition system as well.
\section{Discuss and conclusion}
Due to the low feasibility in the real world, it is not easy for digital adversarial attacks to generate a truly effective threat to computer vision systems. In contrast, physical adversarial attacks are more executability (feasibility) in the real world and can have a significant or even fatal impact on real systems. In recent years, more and more attention has been paid to physical adversarial attacks, and the literature on physical adversarial attacks has been growing every year. However, there are some difficulties with physical adversarial attacks: a) when a small perturbation is implanted, the perturbation is easily ignored by the physical properties of the camera; b) when a large perturbation is implanted, it is easily noticed by humans; c) it is difficult to simulate the experimental setup that generates the perturbation.

From the above analysis, a perfect physical adversarial example not only allows the target model to generate false predictions with a higher confidence, but the perturbations are completely imperceptible to the human eye. From the analysis of Section 3 of this paper, it can be concluded that the current use of optical devices to launch adversarial attack is closer to generating a perfect physical adversarial example, but most experimental setups for such methods are difficult to deploy, so developing a physical adversarial attack where the experimental setup is simple to deploy while the perturbations are imperceptible to the human eye is a direction worth exploring in the future.

\section*{Acknowledgments}
This work was partially supported by the National Natural Science Foundation of China(No. 62171202, No. 62272131), National Science and Technology Major Project Carried on by Shenzhen (CJGJZD20200617103000001), Shenzhen Basic Research Project of China (JCYJ20200109113405927), HKU-SCF FinTech Academy, Shenzhen-Hong Kong-Macao Science and Technology Plan Project (Category C Project: SGDX20210823103537030), and Theme-based Research Scheme of RGC, Hong Kong (T35-710/20-R).



%

\bibliographystyle{IEEEtran}
\bibliography{IEEEexample}

\newpage

\vfill

\end{document}